\documentclass[12pt]{article}
\usepackage{amssymb}
\usepackage{graphicx}
\usepackage{amsmath}
\usepackage{amsfonts}
\usepackage{amssymb}
\usepackage{subfigure}
\usepackage{overpic}
\usepackage{epsfig}
\usepackage[linesnumbered,vlined,slide,ruled]{algorithm2e}

\newcommand{\bb}[1]{\boldsymbol{\mathrm{#1}}}

\def\Tr{\mathrm{T}}

\newcommand{\EE}{\mathbb{E}}
\newcommand{\RR}{\mathbb{R}}
\newcommand{\HH}{\mathbb{H}}

\newcommand{\xx}{\bb{x}}
\newcommand{\yy}{\bb{y}}

\newcommand{\ppp}{\bb{p}}
\newcommand{\qqq}{\bb{q}}

\newcommand{\sign}{\mathrm{sign}}

\newcommand{\vv}{\bb{v}}

\newcommand{\aaa}{\bb{a}}

\newcommand{\Pp}{\bb{P}}

\newcommand{\Kk}{\bb{K}}

\newcommand{\Bb}{\bb{B}}

\newcommand{\Vv}{\bb{V}}

\newcommand{\Ii}{\bb{I}}

\newcommand{\LLambda}{\bb{\Lambda}}
\newcommand{\SSigma}{\bb{\Sigma}}
\newcommand{\aalpha}{\bb{\alpha}}
\newcommand{\bbeta}{\bb{\beta}}

\newcommand\tr{\mathrm{tr}\,}

 {\everymath{\displaystyle\everymath{}}\array}%
 {\endarray}

\begin{document}

\title{Kernel diff-hash}

\author{
Michael M. Bronstein\\
{\small Institute of Computational Science}\\
{\small Faculty of Informatics,}\\ 
{\small Universit{\`a} della Svizzera Italiana}\\ 
{\small Via G. Buffi 13, Lugano 6900, Switzerland}\\
{\tt\small michael.bronstein@usi.ch}\\
}

\maketitle

\begin{abstract}

This paper presents a kernel formulation of the recently introduced diff-hash algorithm for the construction of similarity-sensitive hash functions. Our kernel diff-hash algorithm that shows superior performance on the problem of image feature descriptor matching.

\end{abstract}

\section{Introduction}

Efficient representation of data in compact and convenient way to similarity-sensitive hashing methods, first considered in \cite{Indyk99} and later in \cite{bawa2005lsh,Shakhnarovich05,Kulis09,weiss2009spectral,raginsky-locality}.
Similarity-sensitive hashing methods can be regarded as a particular instance of {\em supervised metric learning} \cite{athitsos04,wang10}, where one tries to construct a hashing function on the data space that preserves known similarity on the training set. Typically, the similarity is binary and can be related to hash collision probability (similar points should collide, and dissimilar points should not collide). Such methods have been enjoying increasing popularity in the computer vision and pattern recognition community  in image analysis and retrieval \cite{jain08_,Torralba08a,jegou2008hamming,jegou2009packing,wang10a,jegou2010product}, video copy detection \cite{videodna}, and shape retrieval \cite{ovsjanikov2009shape}.

Shakhnarovich \cite{Shakhnarovich05} considered parametric hashing functions with affine transformation of the data vectors (projection matrix and threshold vector) followed by the sign function. He posed the problem of similarity-sensitive hash construction as boosted classification, where each dimension of the hash acts as a weak binary classifier. The parameters of the hashing function were learned using AdaBoost. 
In \cite{ldahash}, we used the same setting of the problem and proposed a much simpler algorithm, wherein projections were selected as eigenvectors of the ratio or difference of covariance matrices of similar and dissimilar pairs of data points; the former method was dubbed as LDA-hash and the latter as diff-hash. Applying these methods to SIFT local features in images \cite{Lowe04g}, very compact and accurate binary descriptors were produced. %

The inspiration to this paper is the diff-hash method \cite{ldahash}. While being remarkably simple and efficient, this method suffers from two major limitations. First, the length of the hash is limited by the descriptor dimensionality. In some situations, this is a clear disadvantage, as longer hashes allow to produce more accurate matching. Secondly, the affine hashing functions are in many cases too simple and fail to represent correctly the structure of the data. 
In this paper, we propose a kernel formulation of the diff-hash algorithm which efficiently resolved both problems. We show the performance of the algorithm on the problem of image descriptor matching using the patches dataset from \cite{Winder07} and show that it outperforms the original diff-hash.

\section{Background}

Let $X \subseteq \RR^n$ denote the data space. 
We denote by $\mathcal{P}$ the set of  pairs of similar data points ({\em positives}) and by $\mathcal{N}$ the set of pairs of dissimilar data points ({\em negatives}). 
The problem of similarity-sensitive hashing is to represent the data in a common space $\HH^m = \{ -1, +1\}^m$ of $m$-dimensional binary vectors with the Hamming metric $d_{\HH^m}(a, b) =  \frac{m}{2} - \frac{1}{2} \sum_{i=1}^m a_i b_i$ by means of a map $\xi: X \rightarrow \HH^m$ such that $d_{\HH^m} \circ (\xi \times \xi) |_\mathcal{P} \approx 0$ on and $d_{\HH^m} \circ (\xi \times \xi) |_\mathcal{N} \approx m$.
Alternatively, this can be expressed as having $\EE \{ d_{\HH^m} \circ (\xi \times \eta) | \mathcal {P} \}\approx 0$ (i.e., the hash has high collision probability on the set of positives) and $\EE \{ d_{\HH^m} \circ (\xi \times \eta) | \mathcal {N} \} \gg 0$. 
The former can be interpreted as the {\em false negative rate} (FNR) and the latter as the {\em false positive rate} (FPR).

\subsection{Similarity-sensitive hashing (SSH)}

To further simplify the problem, Shakhnarovich \cite{Shakhnarovich05} considered parametric hashing function of the form $\xi(\xx) = \sign(\Pp \xx + \aaa)$, where $\Pp$ is $m\times n$ {\em projection} matrix and $\aaa$ is an $m\times 1$ {\em threshold} vector. 
The similarity-sensitive hashing (SSH) algorithm considers the hash construction as boosted binary classification, where each hash dimension acts as a weak binary classifier. %
For each dimension, AdaBoost is used to maximize the following loss function
\begin{eqnarray}
\label{eq:mm_loss-ssh}
\min_{\ppp_i, a_i}  \, \sum_{(\xx,\xx') \in \mathcal{P} \cup \mathcal{N}} w_{i}(\xx,\xx') s(\xx,\xx') \xi_i(\xx) \xi_i(\xx'), 
\end{eqnarray}
where $\xi_i(\xx) = \sign(\ppp_i^\Tr \xx + a_i)$, $s(\xx, \xx') = 1$ for $(\xx,\xx') \in \mathcal{N}$ and $0$ for $(\xx,\xx') \in \mathcal{P}$ and $w_{i}(\xx,\xx')$ is the AdaBoost weigh for pair $(\xx,\xx')$ at $i$th iteration. 
Shakhnarovich \cite{Shakhnarovich05} selected $\ppp_i$ as the axis projection onto which minimizes the objective. 
In \cite{videodna,cmssh}, minimization problem~(\ref{eq:mm_loss-ssh}) was relaxed in the following way : First, removing the non-linearity and setting $a_i = 0$, find the projection vector $\ppp_i$. Then, fixing the projection $\ppp_i$, find the threshold $a_i$. 
The disadvantages of the boosting-based SSH is first high computational complexity, and second, the tendency to find unnecessary long hashes.\footnote{The second problem can be partially resolved by using sequential probability testing \cite{waldhash} which creates hashes of minimum expected length.}

\subsection{Diff-hash}

In \cite{ldahash}, we proposed a simpler approach, computing the similarity-sensitive hashing by minimizing
\begin{eqnarray}
\label{eq:mm_loss-lda}
L(\xi) & = & \alpha \EE \{ d_{\HH^m} \circ (\xi \times \xi) | \mathcal {P} \} -  \EE \{ d_{\HH^m} \circ (\xi \times \xi) | \mathcal {N} \} \nonumber \\
&=& {\textstyle \frac{m(\alpha - 1)}{2} }+ {\textstyle\frac{1}{2}} \EE \{ \xi^\Tr \xi | \mathcal{N} \} - {\textstyle\frac{\alpha}{2}} \EE \{ \xi^\Tr \xi | \mathcal{P} \}
\end{eqnarray}
w.r.t. the map $\xi$. Problem~(\ref{eq:mm_loss-lda}) is equivalent, up to constants, to minimizing the correlations 
\begin{eqnarray}
\label{eq:mm_loss-lda1}
L(\Pp,\aaa) & = &  \EE \{ \sign(\Pp \xx + \aaa)^\Tr\sign(\Pp \xx + \aaa) | \mathcal{N} \} \nonumber \\
&-& \alpha \EE \{ \sign(\Pp \xx + \aaa)^\Tr\sign(\Pp \xx + \aaa) | \mathcal{P} \} 
\end{eqnarray}
w.r.t. the projection matrix $\Pp$ and threshold vector $\aaa$. 
The first and second terms in (\ref{eq:mm_loss-lda1}) can be thought of as FPR and FNR, respectively. The parameter $\alpha$ controls the tradeoff between FPR and FNR. The limit case $\alpha \gg 1$ effectively considers only the positive pairs ignoring the negative set. 

Problem (\ref{eq:mm_loss-lda1}) is a highly non-convex non-linear optimization problem difficult to solve straightforwardly. 
Following \cite{videodna,cmssh}, we simplify the problem in the following way. First, ignore the threshold and solve a simplified problem without the sign non-linearity for projection matrix $\Pp$, 
\begin{eqnarray}
\label{eq:diff-hash1}
& \displaystyle\min_{\Pp^\Tr \Pp = \Ii} & \EE \{ (\Pp \xx )^\Tr (\Pp\xx ) | \mathcal{N} \} - \alpha \EE \{ (\Pp \xx )^\Tr (\Pp \xx ) | \mathcal{P} \}  = \nonumber\\
& \displaystyle\min_{\Pp^\Tr \Pp = \Ii} & \tr (\Pp^\Tr \EE \{ \xx \xx^\Tr | \mathcal{N} \} \Pp) - \alpha \tr (\Pp^\Tr \EE \{ \xx \xx^\Tr | \mathcal{P} \} \Pp)=  \nonumber\\
& \displaystyle\min_{\Pp^\Tr \Pp = \Ii} & \tr (\Pp^\Tr (\SSigma_\mathcal{N} - \alpha \SSigma_\mathcal{P}) \Pp), 
\end{eqnarray}
where $\SSigma_\mathcal{P}, \SSigma_\mathcal{N}$ denote the $n \times n$ covariance matrices of the positive and negative data. 
The solution of (\ref{eq:diff-hash1}) is given explicitly as $\Pp = [\lambda^{1/2}_{n-m+1}\vv_{n-m+1}, \hdots , \lambda^{1/2}_n \vv_{n}]^\Tr$, the $m$ smallest eigenvectors of the matrix $\SSigma_\mathcal{N} - \alpha \SSigma_\mathcal{P} = \Vv \LLambda \Vv^\Tr$ of weighted covariance differences.\footnote{The name of the algorithm {\em diff-hash} refers in fact to this covariance difference matrix.}

Second, fixing the projections find optimal threshold vector $\aaa$, 
\begin{eqnarray*}
&\displaystyle\min_{\aaa}& \EE \{ \sign(\Pp \xx + \aaa)^\Tr\sign(\Pp \xx' + \aaa) | \mathcal{N} \} \\
&& - \alpha \EE \{ \sign(\Pp \xx + \aaa)^\Tr\sign(\Pp \xx' + \aaa) | \mathcal{P} \} = \\
&\displaystyle\min_{\{a_i\}}& \textstyle\sum_{i=1}^m\EE \{ \sign(\ppp_i^\Tr \xx + a_i) \sign(\ppp_i^\Tr \xx + a_i) | \mathcal {N} \} \\
&& - \alpha \textstyle\sum_{i=1}^m \EE \{ \sign(\ppp_i^\Tr \xx + a_i) \sign(\ppp_i^\Tr \xx + a_i) | \mathcal {P} \}.
\end{eqnarray*}
The problem is separable and can be solved independently in each dimension $i$. 
The above terms are the false positive and negative rates as function of the threshold $a_i$, 
\begin{eqnarray*}
\label{eq:prob1}
\mathrm{FNR}(a_i) &=& \mathrm{Pr}(\ppp_i^\Tr \xx  + a_i < 0 \,\,\, \mathrm{and}\,\,\, \ppp_i^\Tr \xx'  + a_i > 0 | \mathcal{P})\\
&+& \mathrm{Pr}(\ppp_i^\Tr \xx  + a_i > 0 \,\,\, \mathrm{and}\,\,\, \ppp_i^\Tr \xx'  + a_i < 0 | \mathcal{P}) \nonumber 
\end{eqnarray*}
and 
\begin{eqnarray*}
\label{eq:prob2}
\mathrm{FPR}(a_i) &=& \mathrm{Pr}(\ppp_i^\Tr \xx + a_i < 0 \,\,\, \mathrm{and}\,\,\, \ppp_i^\Tr \xx' + a_i < 0 | \mathcal{N})  \\
&+& \mathrm{Pr}(\ppp_i^\Tr \xx + a_i > 0 \,\,\, \mathrm{and}\,\,\, \ppp_i^\Tr \xx' + a_i > 0 | \mathcal{N}). \nonumber 
\end{eqnarray*}
The above probabilities 
can be estimated from histograms (cumulative distributions) of $ \ppp_i^\Tr \xx$ and $ \qqq_i^\Tr \yy$ on the positive and negative sets. 
The optimal threshold 
\begin{eqnarray}
\label{eq:opt_thr}
a_i^* &=& \mathop{\mathrm{argmin}}_{a} \,\, \alpha \mathrm{FNR}(a) + \mathrm{FPR}(a)
\end{eqnarray}
is obtained by means of one-dimensional exhaustive search.

\section{Kernel diff-hash}

An obvious disadvantage of diff-hash (and spectral methods in general) compared to AdaBoost-based methods is that it must be {\em dimensionality-reducing}: since we compute projection $\Pp$ as the eigenvectors of a covariance matrix of size $n\times n$, the dimensionality of the embedding space must be $m \leq n$. 
This restriction is limiting in many cases, as first it depends on the data dimensionality, and second, such a dimensionality may be too low and a longer hash would achieve better performance. 
Furthermore, the affine parametric form of the embedding $\xi$ is in many cases an oversimplification, and some more generic map is required. 
%

In this paper, we cope with both problems using a kernel formulation, which transforms the data into some feature space that is never dealt with explicitly (only inner products in this space, referred to as {\em kernel} \cite{kernel}, are required). 
In order to simplify the following discussion, since the problem is separable (as we have seen, projection in each dimension corresponds to a eigenvector of the covariance matrix difference), we consider one-dimensional projections.
The whole method is summarized in Algorithm~\ref{alg:2}.

\subsection{Projection computation}

Let $k_X : X\times X \rightarrow \RR$ be a positive semi-definite kernel, and let $\phi: \xx \mapsto k_X(\cdot, \xx)$. Thus, $\phi$ maps the data into some feature space, which we represent here as a Hilbert space $\mathcal{V}$ (possibly of infinite dimension) with an inner  product $\langle \cdot, \cdot \rangle_\mathcal{V}$, and satisfies $k_X(\xx,\xx') = \langle k_X(\cdot, \xx), k_X(\cdot, \xx') \rangle_{\mathcal{V}} = \langle \phi(\xx), \phi(\xx') \rangle_{\mathcal{V}}$. 

The idea of kernelization is to replace the original data $X$ with the corresponding feature vectors $\phi(X)$, replacing the linear projection $\ppp^\Tr \xx$ with $p(\xx) = \sum_{i=1}^l \beta_i \langle \phi(\xx_i), \phi(\xx) \rangle_{\mathcal{V}} =  \bbeta^\Tr [ k_X(\xx_1, \xx) \hdots k_X(\xx_l, \xx)]$. Here, $\bbeta$ is a vector of unknown linear combination coefficients, and $\xx_1,\hdots, \xx_l$ denote some representative points in the data space. 

In this formulation, at the projection computation stage we minimize, for each dimension 
\begin{eqnarray*}
& \displaystyle\min_{\bbeta}& \frac{1}{|\mathcal{N}|} \sum_{(\xx,\yy)\in \mathcal{N}} p(\xx)  q(\yy) - \frac{\alpha}{|\mathcal{P}|} \sum_{(\xx,\yy)\in \mathcal{P}} p(\xx)  q(\yy) = \\
&\displaystyle\min_{\bbeta}& \frac{1}{|\mathcal{N}|} \sum_{(\xx,\xx')\in \mathcal{N}}  \sum_{i,j=1}^l \beta_i \langle \phi(\xx_i), \phi(\xx) \rangle_{\mathcal{V}}  \beta_j \langle \phi(\xx_j), \phi(\xx') \rangle_{\mathcal{V}} = \\
&&-\frac{\alpha}{|\mathcal{P}|} \sum_{(\xx,\xx')\in \mathcal{P}}  \sum_{i,j=1}^l \beta_i \langle \phi(\xx_i), \phi(\xx) \rangle_{\mathcal{V}} \beta_j \langle \phi(\xx_j), \phi(\xx') \rangle_{\mathcal{V}} = \\
&\displaystyle\min_{\bbeta}&  \frac{1}{|\mathcal{N}|} \bbeta^\Tr \Kk_\mathcal{N}  \Kk^\Tr_\mathcal{N} \bbeta -\frac{\alpha}{|\mathcal{P}|} \bbeta^\Tr \Kk_\mathcal{P}  \Kk^\Tr_\mathcal{P} \bbeta 
= \displaystyle\min_{\bbeta}  \bbeta^\Tr \Kk \bbeta, 
\end{eqnarray*}
where $\Kk_\mathcal{N}$ and $\Kk_\mathcal{P}$ denote $l\times |\mathcal{N}|$ and $l\times |\mathcal{P}|$ matrices with elements $k_X(\xx_i,\xx)$. 
The optimal projection coefficients $\aalpha$ minimizing are given as the smallest eigenvectors of the $l\times l$ matrix $\Kk = \frac{1}{|\mathcal{N}|} \Kk_\mathcal{N}  \Kk^\Tr_\mathcal{N} - \frac{\alpha}{|\mathcal{P}|} \Kk_\mathcal{P}  \Kk^\Tr_\mathcal{P} $.

The kernel $k_X$ can be selected to account correctly for the structure of the data space $X$. In our formulation, the dimensionality of the hash is bounded by the number of the basis vectors,  $m  \leq l$, which is limited only by the training set size and computational complexity.

\subsection{Threshold selection}

As previously, the threshold should be selected to minimize the false positive and false negative rates, that can be expressed, as previously, as 
\begin{eqnarray*}
\label{eq:fn2}
\mathrm{FNR}(a) &=& 
\mathrm{Pr}(p(\xx) + a < 0 \,\,\, \mathrm{and} \,\,\, p(\xx') + a > 0 | \mathcal{P})  \\
&+& \mathrm{Pr}(p(\xx) + a > 0 \,\,\, \mathrm{and} \,\,\, p(\xx') + a < 0 | \mathcal{P}),\nonumber \\
\mathrm{FPR}(a) &=& 
\mathrm{Pr}(p(\xx) + a < 0 \,\,\, \mathrm{and} \,\,\, p(\xx') + a < 0  | \mathcal{N}) \\
&+& \mathrm{Pr}(p(\xx) + a > 0 \,\,\, \mathrm{and} \,\,\, p(\xx') + a > 0  | \mathcal{N}), \nonumber
\end{eqnarray*}
The optimal threshold is  obtained as 
\begin{eqnarray}
\label{eq:opt_thr2}
a^* &=& \mathop{\mathrm{argmin}}_{a} \,\,\,\alpha \mathrm{FNR}(a) +  \mathrm{FPR}(a).
\end{eqnarray}

%

\subsection{Hash function application}

Once the coefficients $\Bb$ and threshold $\aaa$ are computed, given a new data point $\xx$, the corresponding $m$-dimensional binary hash vector is constructed as 
$\xi(\xx) = \sign(\Bb (k_X(\xx_1,\xx),\hdots,k_X(\xx_l,\xx))^\Tr + \aaa)$.
Note that this embedding is kernel-dependent and has a more generic form than the affine transformation used in \cite{Shakhnarovich05,ldahash}.

\begin{algorithm}[t!]
\KwIn{Positives set $\mathcal{P} \subset X\times X$, Negatives set $\mathcal{N} \subset X\times X$; 
Dimensionality of the hash $m$; Kernel $k_X$; Set of vectors $\xx_1,\hdots,\xx_l$.}
\KwOut{Optimal combination coefficient matrix $\Bb$ of size $m\times l$; optimal offset vector $\aaa$ of size $m\times 1$.}

Compute the kernel matrices $\Kk_\mathcal{P}, \Kk_\mathcal{N}$ of size $l\times |\mathcal{P}|$  and $l\times |\mathcal{N}|$, respectively.

Compute the matrix $\Kk = \frac{1}{|\mathcal{N}|} \Kk_\mathcal{N}  \Kk^\Tr_\mathcal{N} - \frac{\alpha}{|\mathcal{P}|} \Kk_\mathcal{P}  \Kk^\Tr_\mathcal{P} $.

Perform eigendecomposition  $\Kk = \Vv \LLambda \Vv^\Tr$. 

\For{$i=1,\dots,m$}{

Set the $i$th row of the coefficient matrices to be the $i$th smallest eigenvectors, $\bbeta^\Tr_i = \lambda_{n-1+1}\vv^\Tr_{n-i+1}$.

Compute the projection $p_i(\xx) = \bbeta_i^\Tr \Kk_X$.

Compute the rates $\mathrm{FNR}(a_i)$ and $\mathrm{FPR}(a_i)$ for $p_i(\xx)+a_i$, as function of threshold $a_i$. 

Compute the optimal thresholds 
\begin{eqnarray*}
a_i^* &=& \mathop{\mathrm{argmin}}_{a} \alpha \mathrm{FNR}(a) + \mathrm{FPR}(a).
\end{eqnarray*}

}

\caption{Kernel diff-hash algorithm. \label{alg:2}}
\end{algorithm}


\section{Results}


\begin{figure}[t!]
    \begin{center}
\includegraphics[width=1\linewidth]{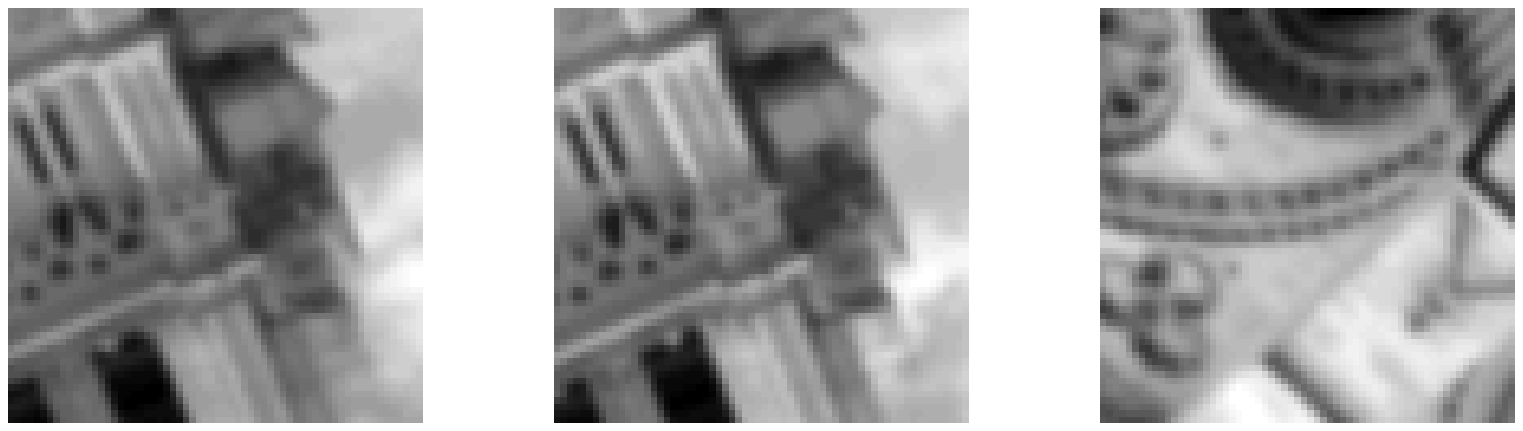} \vspace{-10mm}\\
\includegraphics[width=1\linewidth]{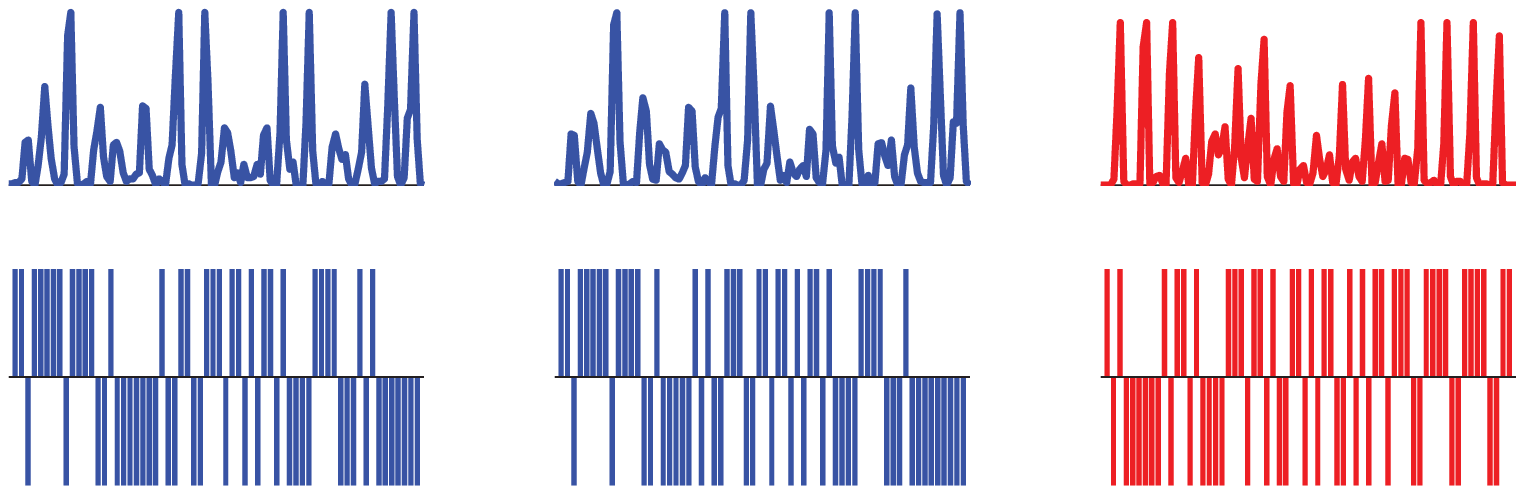}\\
   \caption{\label{fig:pairs} \small Example of a positive (left, middle) and negative (left, right) pair of image patches and corresponding descriptors. First row: patches, second row: SIFT descriptors, third row: binary descriptors of length $32$ produced using kDIF.}
    \end{center}
\end{figure}

In order to test our approach, we applied it to the problem of image feature matching. This problem is a core of many modern Internet-scale computer vision applications, including city scale reconstruction \cite{Agarwal09}. 
The basic underlying task in these problems, repeated millions and billions of times, is the comparison of local image features  (SIFT \cite{Lowe04g} or similar methods \cite{Miko05,Bay08,Tola10}). Typically, these features are represented by means of multidimensional descriptors vectors (e.g. SIFT is $128$-dimensional) and compared using the Euclidean distance. 
With very large datasets (containing $10^6-10^9$ feature points), severe scalability issues are encountered, including problems of storage and similarity query on feature descriptors. 
Efficient representation and comparison of feature descriptors have been addressed in many recent works in the computer vision community (see, e.g., \cite{Miko03,mikolajczyk2007improving,Tuytelaars07,Hua07,Winder07,Winder09,Chandrasekhar09,Brown10}). 
In \cite{ldahash}, we proposed using similarity-sensitive hashing methods to produce compact {\em binary descriptors} \cite{ldahash}. Such descriptors have several appealing properties that make them especially suitable in large-scale applications. First, they are compact (typically, $64-256$ bits, compared to at least $1024$ required for the standard SIFT) and easy to store in standard databases. Second, the comparison of binary descriptors is done using the Hamming metric, which amounts to XOR and bit count -- an operation that can be carried out extremely efficiently on modern CPU architectures, significantly faster than the computation of Euclidean or other $L_p$ distances. 
Finally, the construction of the binarization transformations involves metric learning, thus modeling more correctly the distance between the descriptors, which is usually non-Euclidean. In particular, this allows to compensate for imperfect invariance of the descriptor (since viewpoint transformations are only approximately locally affine) and cope with descriptor variability in pairs of images with wide baseline. As a result of this last property, the use of similarity-sensitive hashing reduces the descriptor size while actually {\em improving} its performance \cite{ldahash}, unlike other methods that typically come at the price of decreased performance.

In our experiments, we used data from \cite{Winder07}. The datasets contained rectified and normalized $64\times 64$ patches extracted from multiple images depicting three different scenes (Trevi fountain, Notre Dame cathedral, and Half Dome). The first two scenes were similar representing architectural landmarks; the last scene was different representing a natural mountain environment. In each scene, a total of nearly $100$K patches corresponding to around $30$K different feature points were available; each feature appeared multiple times. 
For training, we used $100$K pairs of patches corresponding to different views of the same points as positives, and $200$K pairs of patches from different points as negatives (Figure~\ref{fig:pairs}). 
For testing, a different subset of the dataset containing $50$K positive and $50$K negative pairs was used.

In each patch, a $128$-dimensional ($8$-bit per dimension) SIFT descriptor was computed using the toolbox of Vedaldi \cite{Vedaldi07}. 
We compared the performance of binary descriptor obtained by means of the diff-hash method of Strecha {\em at al.} \cite{ldahash} (DIF) and our kernel version (kDIF). Diff-hash appeared to be the best performing algorithm in an extensive set of evaluations done in \cite{ldahash}. Since kDIF is an extended version of DIF, we choose to compare to this method.  
In both methods, we used the value $\alpha=25$ which was experimentally found to produce the best results. In kDIF, we used a Gaussian kernel with the Mahalanobis distance of the form $k_X(\xx,\xx') = \exp\{ -(\xx - \xx')^\Tr \SSigma_X^{-1/2}(\xx - \xx') \}$. The same training and testing data were used for all methods. 
For reference, we show the Euclidean distance between the original SIFT descriptors. 

%

\begin{figure*}[h!]
    \begin{center}
\includegraphics[width=0.75\linewidth]{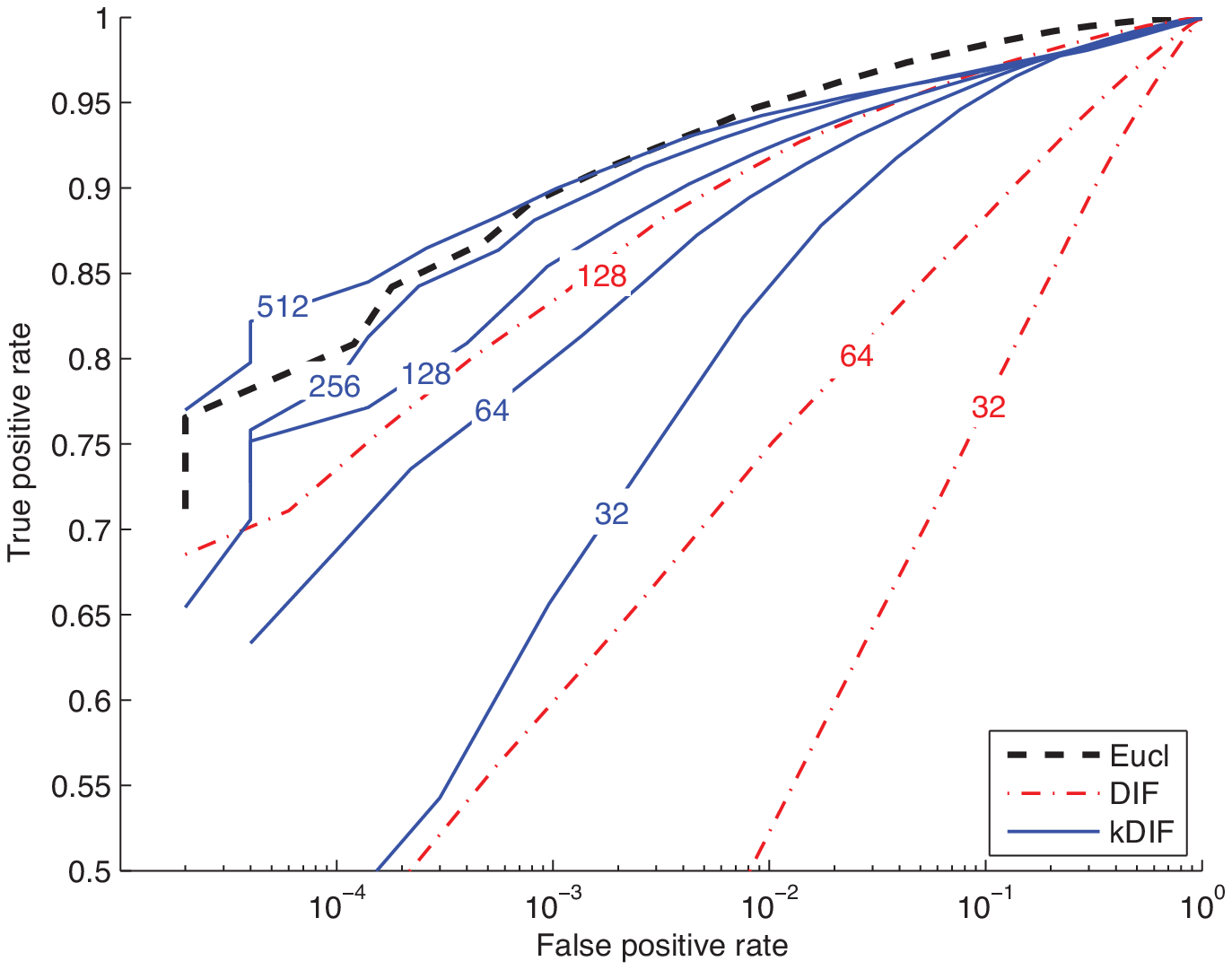}\\
\small (a) halfdome-halfdome \vspace{2mm}\\
\includegraphics[width=0.75\linewidth]{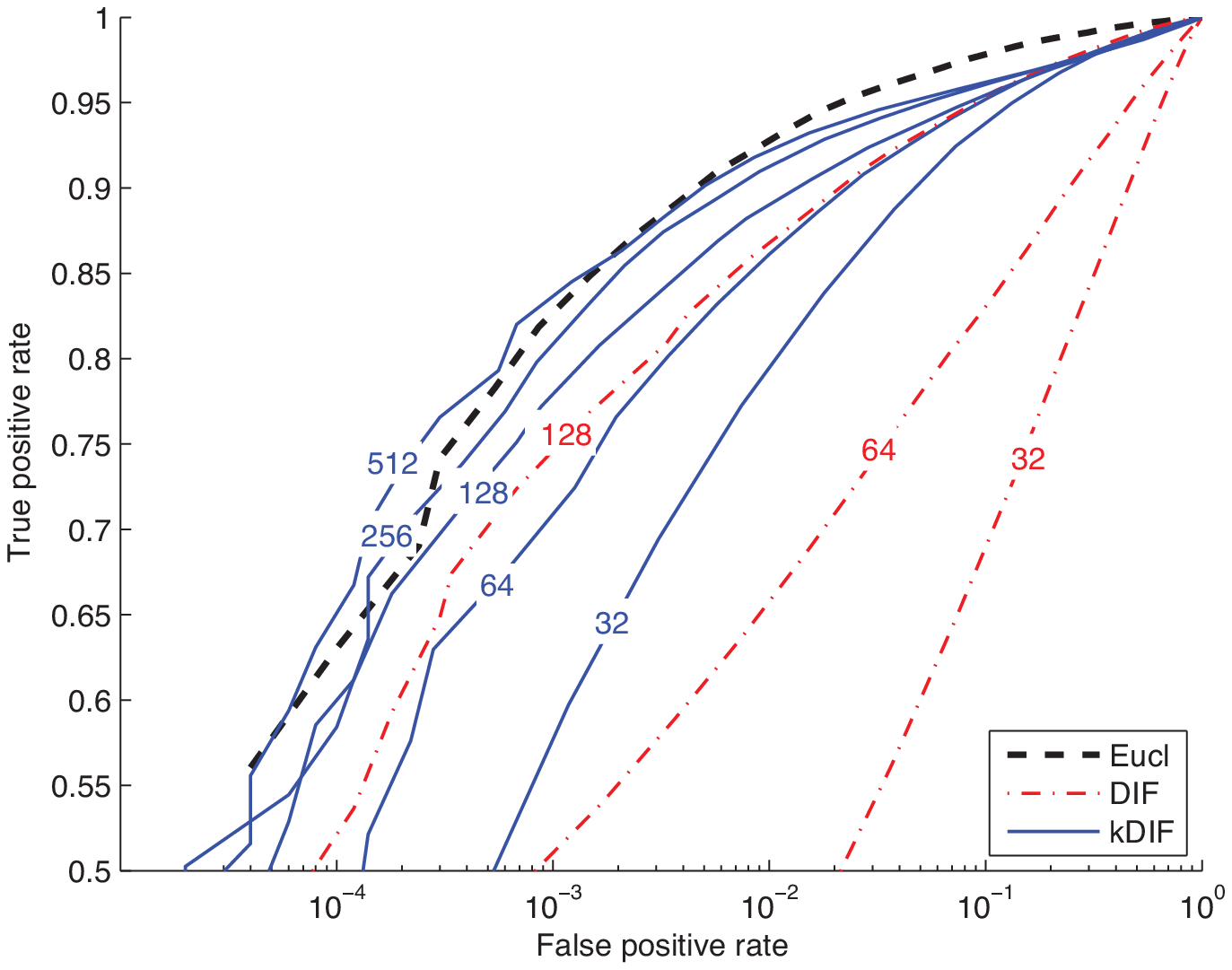}\\
\small (b) halfdome-notredame \\
   \caption{\label{fig:roc1} \small ROC curves showing the performance of Euclidean distance between SIFT descriptors (dashed black) and Hamming distance between binary vectors of different dimension $m = 32, 64, \hdots, 512$ constructed using DIF (dash-dot red) and kDIF (solid blue) hashing algorithms. Captions follow the convention {\em training-test}. }
    \end{center}
\end{figure*}

\begin{figure*}[h!]
    \begin{center}
\includegraphics[width=0.75\linewidth]{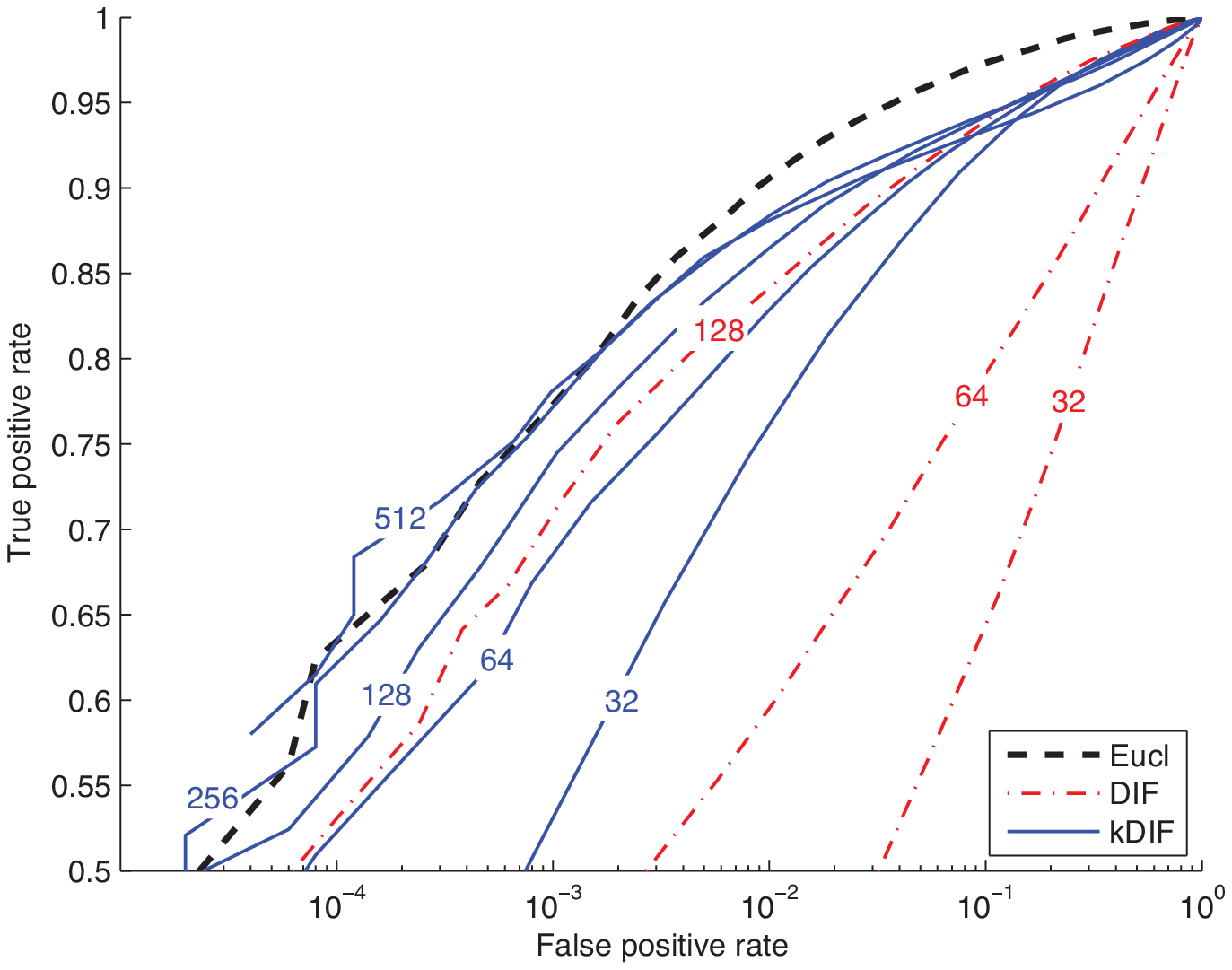}\\
\small  (a) trevi-trevi \\
\includegraphics[width=0.75\linewidth]{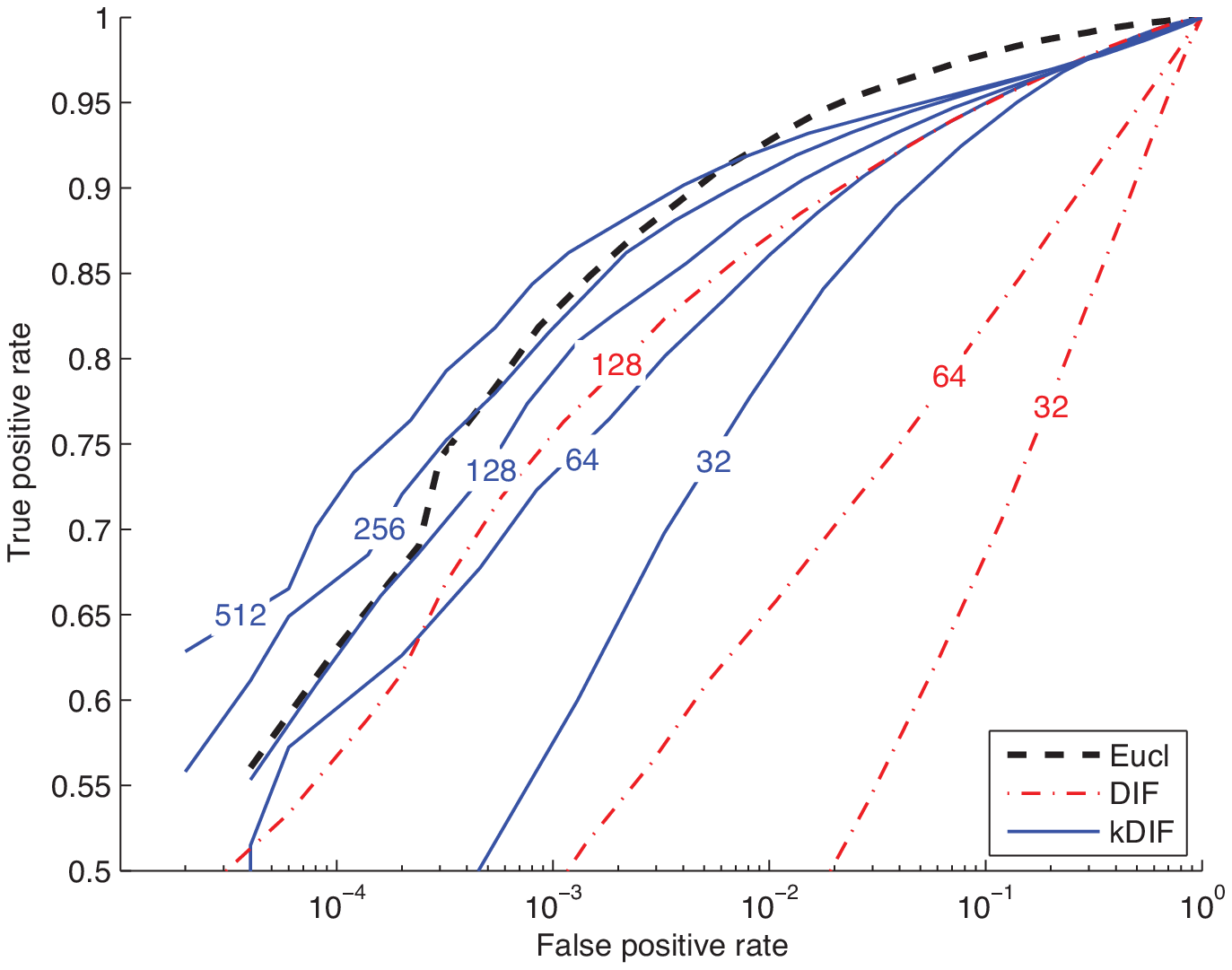}\\
\small  (b) trevi-notredame\\
   \caption{\label{fig:roc2} \small ROC curves showing the performance of Euclidean distance between SIFT descriptors (dashed black) and Hamming distance between binary vectors of different dimension $m = 32, 64, \hdots, 512$ constructed using DIF (dash-dot red) and kDIF (solid blue) hashing algorithms. Captions follow the convention {\em training-test}. }
    \end{center}
\end{figure*}

Figures~\ref{fig:roc1}--\ref{fig:roc2} show the performance of different hashing algorithms as a function of $m$ on different datasets. Several conclusions can be drawn from this figure. First, kDIF appears to consistently outperform DIF on all three scenes for the same hash length $m$. Second, for sufficiently large $m$, our method outperforms SIFT while still being more compact. Third, the learned hashing functions generalize gracefully to other scenes, though slight performance degradation is noticeable when training on mountain scene (Half Dome) and using the learned hash in an architectural scene (Note Dame).

\begin{figure*}[h!]
    \begin{center}
\includegraphics[width=0.75\linewidth]{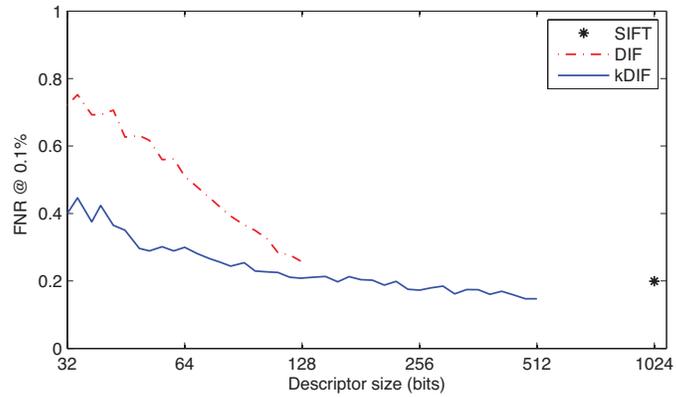} \vspace{2mm}\\
\includegraphics[width=0.75\linewidth]{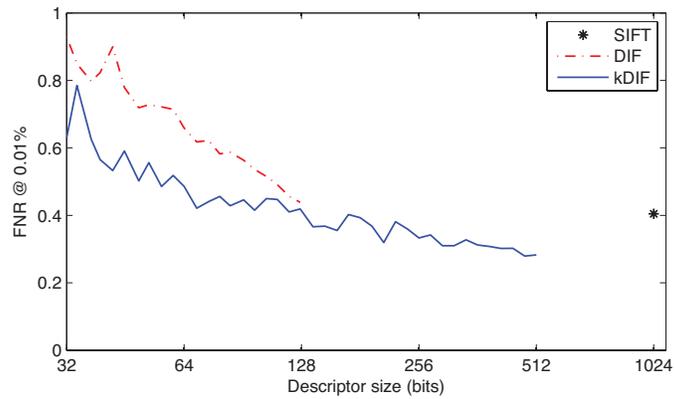}\\
   \caption{\label{fig:size} \small Performance (FNR at $0.1\%$ and $0.01\%$ FPR; the smaller the better) of different methods as function of descriptor size in bits. Training was done on trevi dataset; testing on notredame dataset.}
    \end{center}
\end{figure*}

Figure~\ref{fig:size} compares the performance of different descriptors in terms of FNR at two low FPR points ($0.1\%$ and $0.01\%$). Binary descriptors outperform raw SIFT while being 2-4 more compact (to say nothing about the lower computational complexity of the Hamming distance compared to the Euclidean distance). Second, kDIF consistently outperforms DIF. Third, one can see that using longer hash ($m>128$) increases the performance.

\begin{figure*}[tpb]
    \begin{center}
\includegraphics[width=1\linewidth]{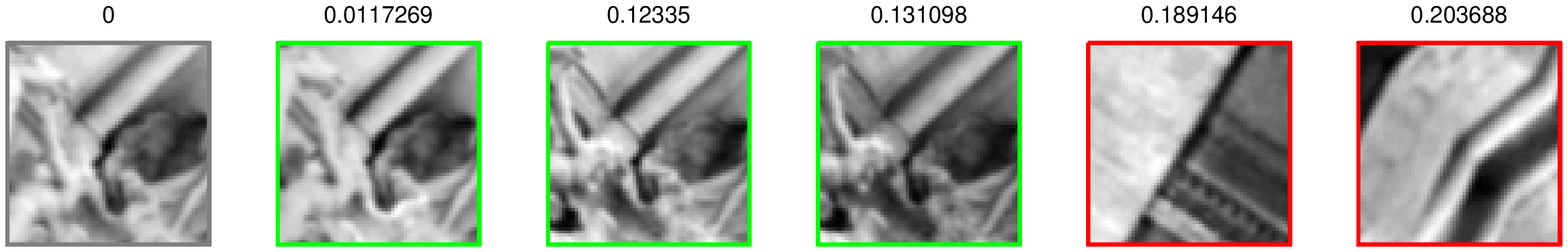}\\
\includegraphics[width=1\linewidth]{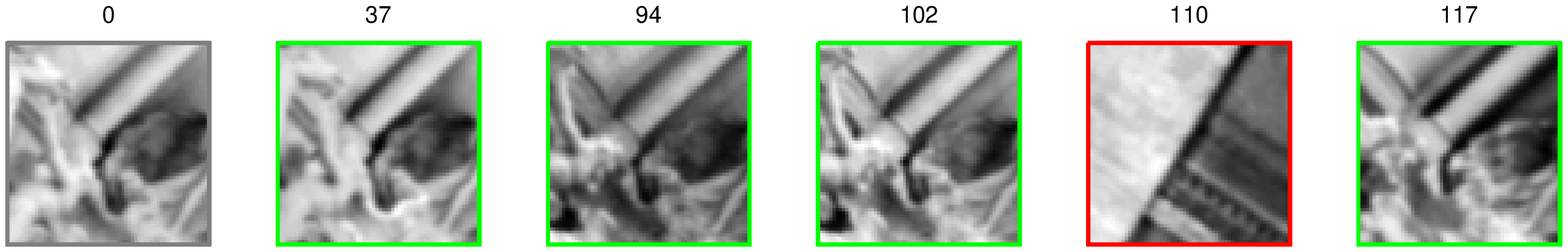}\vspace{3mm}\\
\includegraphics[width=1\linewidth]{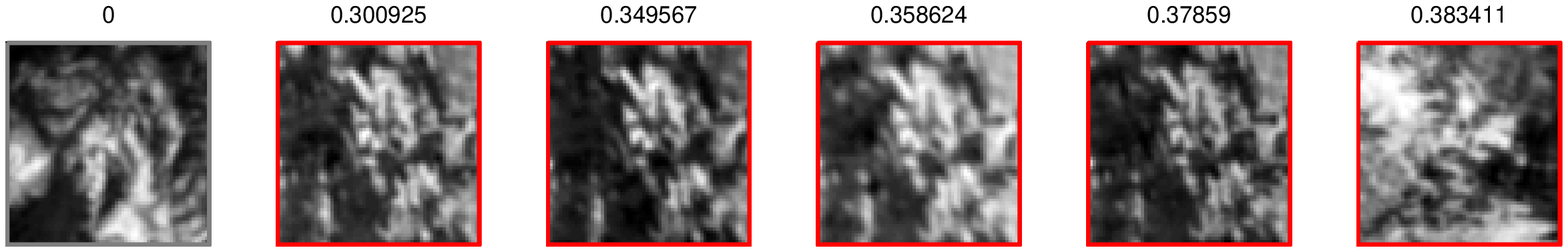}\\
\includegraphics[width=1\linewidth]{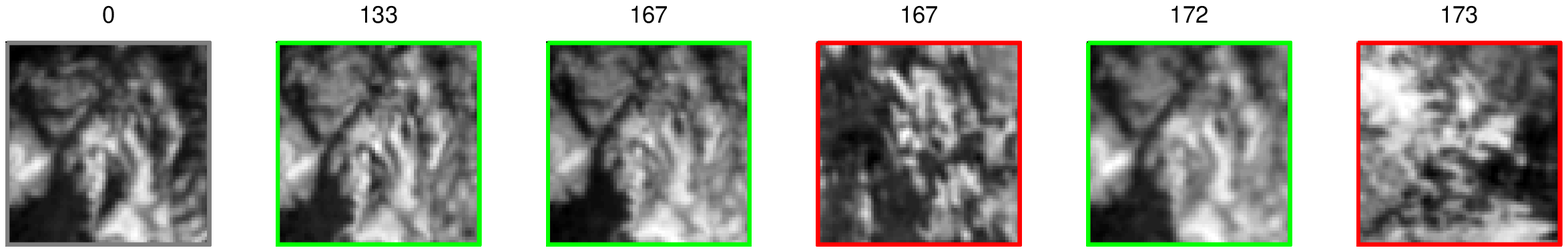}\vspace{3mm}\\
\includegraphics[width=1\linewidth]{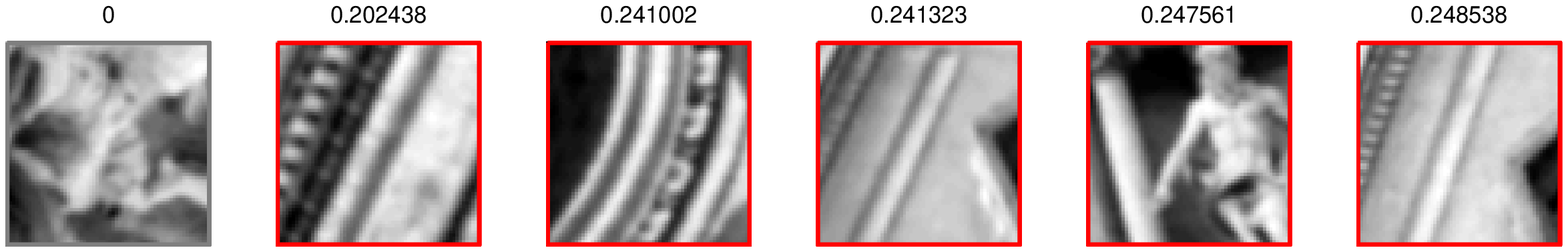}\\
\includegraphics[width=1\linewidth]{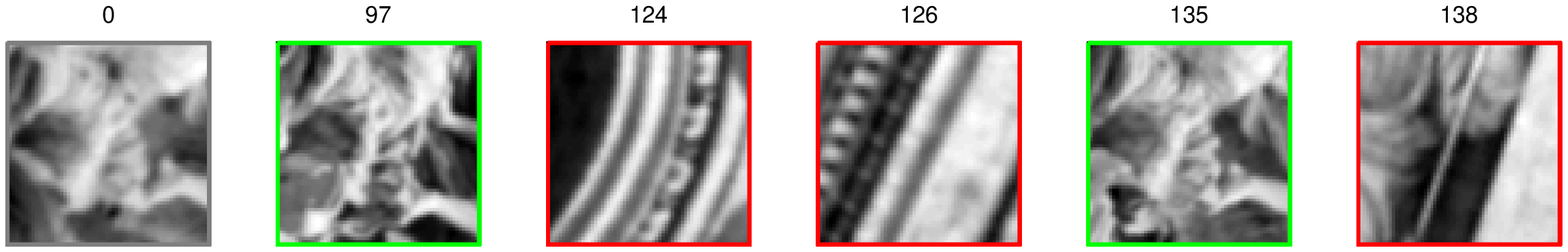}\vspace{3mm}\\
   \caption{\label{fig:matches} \small First matches using Euclidean distance between SIFT descriptors (odd rows) and Hamming distance between $512$-dimensional binary vectors constructed using our kDIF hashing algorithms (even rows). Query image is shown on the left, first five matches are shown on the right. Numbers indicate the distance from query. Wrong matches are marked in red, correct matches are marked in green. }
    \end{center}
\end{figure*}

Figure~\ref{fig:matches} shows a few examples of first matches between patch descriptors obtained using Euclidean distance and the Hamming distance on the hashed descriptors using our method. Our method provides superior performance.


\section{Conclusions}

We presented kernel formulation of diff-hash similarity-sensitive hashing algorithm and showed how this method can be used to produce efficient and compact binary feature descriptors. Though we showed results with SIFT, the method is generic and can be applied to any local feature descriptor. Our method showed superior results compared to the original diff-hash proposed in \cite{ldahash}, and is more generic as it allows to obtain hashes of any length and also incorporate nonlinearity through the choice of the kernel.

\small
\bibliographystyle{plain}
\bibliography{ldahash,string,vision,misc,new}

\end{document}